# PPG-Based Affect Recognition with Long-Range Deep Models: A Measurement-Driven Comparison of CNN, Transformer, and Mamba Architectures

Karim Alghoul
School of Electrical Engineering and Computer Science
University of Ottawa
Ottawa, Canada
Karim.Alghoul@uottawa.ca

Hussein Al Osman
School of Electrical Engineering and Computer Science
University of Ottawa
Ottawa, Canada
Hussein.Alosman@uottawa.ca

Abdulmotaleb El Saddik
School of Electrical Engineering and Computer Science - University of Ottawa, Canada
elsaddik@uottawa.ca

*Abstract*—Photoplethysmography (PPG) is increasingly used in wearable affective computing due to its low cost and ease of integration into consumer devices. Recent advances in deep learning have introduced long-range sequence models, such as Transformers, and state-space models, like Mamba, which have demonstrated strong performance on natural language and general time-series tasks. However, it remains unclear whether these architectures offer tangible benefits over widely used Convolutional Neural Networks (CNNs) and Long Short-Term Memory (LSTMs) for PPG-based affect recognition, given that datasets are typically small and noisy. This work presents a measurement-driven comparison of four deep learning architectures, CNN, CNN–LSTM hybrid, Transformers, and Mamba, for classifying arousal, valence, and relaxation states from wrist-based PPG signals. All models are evaluated under a subject-independent 5-fold cross-validation protocol using identical preprocessing, segmentation, and training pipelines. Our results show that the Transformer and Mamba models achieve performance comparable to that of a CNN baseline, but do not consistently outperform it across all tasks. CNNs remain the most effective overall, providing the highest accuracy with the smallest model size, whereas Transformers have a better balance of F1 scores for Arousal and Relaxation. The study provides the first evaluation of Transformer and Mamba models for PPG-based affect recognition, offering practical guidance on model selection for wearable affective monitoring systems.

## I. INTRODUCTION

Advances in signal analysis for speech and facial expressions have been essential in the field of affective computing [1]. Machine learning has significantly contributed to this progress, with algorithms developed to detect emotional states from speech, facial expressions, and gestures [2][3]. Among these methods, neural network approaches are notable for automatically extracting emotional features from speech and facial cues, enabling a deeper understanding of emotional signals [4]. Many applications have been introduced in different fields, ranging from emotion-aware tutoring systems to healthcare, where measuring users' emotions can help nurses identify distressed patients who need immediate intervention [5].

More recently, the focus has expanded from speech and facial expressions to physiological measurements, such as electrocardiograms (ECG) and PPG, driven by the widespread use of wearable devices like sports chest belts, smartphones, and smartwatches [6]. ECG and PPG sensors enable non-invasive, continuous acquisition of physiological signals [7]. These signals provide objective and dependable indicators of affective states, as they are less influenced by social masking than observable cues such as facial expressions or voice tone [8][9][10].

Currently, most commercial smartwatches have built-in PPG sensors, mainly used to measure heart rate. This has expanded the range of applications for PPG sensors beyond simple physical activity tracking [11]. However, since PPG signals are very sensitive to motion and the user's wrist is constantly moving, it has been and still is difficult to develop PPG-based models that are both accurate and able to generalize well for assessing users' affective states [12]. Additionally, existing datasets are limited in participant numbers and were gathered under restricted protocols [13][14]. Affective responses differ among individuals, adding more variability. These limitations make it unclear whether more complex deep learning models can reliably generalize in this setting.

Over the past few years, CNN-based models have been the most widely used architectures for PPG-based affect recognition [15][16][17][13]. More recently, long-range sequence models such as Transformers have become the state-of-the-art (SOTA) in many fields, particularly in natural language processing (NLP) [18]. In parallel, state-space models such as Mamba have emerged as an alternative to Transformers for efficient long-sequence modeling [19]. Despite their success in other domains, it remains unclear whether Transformers and Mamba actually provide added value over CNN-based architectures for PPG-based affect recognition, especially under real-world constraints on wearable data.

**Objectives and Contributions**

This paper investigates the benefit of long-range models, specifically Transformer and Mamba, over CNN-based models for PPG-based affect recognition. We perform a measurement-driven comparison of four deep learning architectures for classifying arousal, valence, and relaxation from wrist-based PPG signals using the WARM-VR dataset [20]. The study is designed to provide practical guidance for model selection in wearable affective computing, highlighting scenarios where it may still be advantageous to deploy classical architectures rather than more complex long-range models.

## II. RELATED WORK

PPG is often used as one component within multimodal affect recognition systems [21], but its use as a single input modality has been more limited due to its vulnerability to motion artifacts and noise [12]. However, recent work indicates that unimodal PPG, despite being less commonly adopted than multimodal pipelines, can achieve good results [15][17][22]. Compared with other physiological measurements, PPG offers several practical advantages that make it well-suited for real-world deployment. It can be acquired unobtrusively via compact, inexpensive wearables (e.g., smartwatches or smartphones), avoiding bulkier instrumentation such as ECG chest straps. Wrist-based PPG sensing is a non-invasive, user-friendly, and cost-effective method that enables continuous monitoring outside laboratory settings [15].

Recent work has explored a range of deep models for PPG-based emotion recognition, with CNNs featuring prominently. Lee M. et al. [15] targeted short-term recognition from a single-beat PPG and used a 1D-CNN for end-to-end feature learning and classification, validating on the DEAP dataset [23]. A related study [17] designed a model that fuses deep features extracted by two CNNs with statistical features extracted from NN intervals. For stress detection, [16] proposed a hybrid CNN (H-CNN) that fuses hand-crafted descriptors with learned features, using wrist PPG from the WESAD dataset [14], which differs from DEAP's valence/arousal formulation by employing a multi-label protocol. Beyond CNNs, LSTM architectures have also been investigated. For PPG specifically, CNN–LSTM hybrids have been effective as well, with a 1D-CNN plus LSTM pipeline applied to contact [22] and non-contact [24] PPG affect recognition. Temporal Convolutional Networks (TCNs) have also been combined with CNNs, improving generalization in PPG-based affect measurement [22].

Long-range sequence modeling has recently gained attention in physiological time series. Transformers, which are effective at capturing global dependencies via self-attention, have been adapted from NLP and Computer Vision (CV) fields to biosignals. For instance, [25] introduced a modified Transformer-based model for affect recognition from multi-channel EEG signals, adeptly accommodating multiple channels simultaneously to distinguish intricate spatiotemporal dynamics associated with affective states and achieving high accuracies in valence and arousal prediction. Another study [26] used a Transformer-based self-supervised learning approach on ECG signals, to estimate the affective state. For PPG, Transformer models have shown utility in tasks such as arrhythmia detection [27] and continuous blood-pressure estimation [28], where longer temporal context is beneficial and training data is relatively abundant, as in clinical or operating-room settings.

In parallel, state-space models such as Mamba present several strengths and are trending as an alternative to Transformer-based models for certain tasks. The trend towards Mamba as an alternative stems from its ability to offer a compelling balance between computational efficiency and the capacity to model long-range dependencies, particularly in applications requiring real-time processing and deployment on devices with limited resources, where Transformers often face challenges due to their quadratic complexity. Lately, many researchers have started using Mamba-based sequence models with physiological signals. For ECG signals, Mamba-based architectures have been used for cardiac abnormality classification, demonstrating efficient long-sequence modeling and competitive accuracy for cardiovascular monitoring [29][19][30]. Other works used Mamba variants with remote PPG (rPPG) to better capture pulse dynamics under challenging lighting and motion [31][32], and for PPG-based blood-pressure estimation via Mamba-UNet blocks that model long-range temporal and vascular variability [33]. For emotion, Mamba-VA combined masked autoencoders, TCNs, and Mamba to predict continuous valence–arousal from video [34].

Despite their proven superiority in other domains, to the best of our knowledge, no prior evaluation of Transformer or Mamba architectures for contact PPG-based affect recognition has been conducted. This gap motivates our study to compare CNN-based baseline models with Transformer and Mamba architectures using raw PPG signals from wearables to estimate arousal, valence, and relaxation.

### III. METHODOLOGY

Fig. 1 shows the overall workflow of the machine learning validation pipeline used in this study.

#### A. Model Architectures Evaluated

We used CNN alone and CNN-LSTM as baseline models, which have been used before on raw PPG signals. We then implemented a Transformer and Mamba models to compare the results.

**CNN model**: The CNN baseline is based on [22] and consists of a three-layer 1D convolutional feature extractor followed by a fully connected classification head. The first convolutional block applies an 8-filter Conv1D layer (*kernel size* = 64, *stride* = 4) using *ReLU* activation and *same* padding. The *max pooling*, *batch normalization*, and 50% *dropout* were applied for regularization. The second convolutional layer increases the channel depth to 16 filters with a smaller *kernel size* and *stride* (32 and 2), again followed by *max pooling*, *normalization*, and *dropout*. A third convolutional block with 8 filters (*kernel size* = 16, *stride* = 1) was added to further refine local temporal features. After flattening, a final fully connected layer with *softmax* activation performs binary classification (high or low).

**CNN-LSTM model**: We implemented a hybrid CNN–LSTM architecture where the model begins with two convolutional blocks identical in structure to the CNN baseline. These layers extract hierarchical temporal features while providing robustness to noise and motion artifacts. The resulting feature maps are then fed into an LSTM layer with 12 *units*, which captures sequential dependencies across the windowed PPG segment. A final fully connected *softmax* layer performs the binary classification of the affective state.

**Transformer model:** The Transformer architecture adapts an encoder-style design to 1D PPG segments by first converting the raw signal into a sequence of patch embeddings. Given an input segment, a 1D convolutional patch-embedding layer (*kernel size* 32, *stride* 32) projects a non-overlapping patch into a latent vector of dimension *$d_{model}$*. We evaluate two configurations of model dimensionality: a lightweight version "Transformer" with *$d_{model}$* =32 and 4 attention *heads*, and a higher-capacity version "Transformer-v2" with *$d_{model}$* =64 and 2 *heads*. To preserve temporal structure, a learned positional embedding is added to each token. The embedded sequence is then passed through a Transformer encoder block (*depth* = 1), consisting of multi-head self-attention, residual connections, and layer normalization, followed by a feed-forward network with GELU activation and an expansion ratio of 2, wrapped in *normalization* and *residual connections*. A 30% *dropout* is applied to both attention and feed-forward outputs for regularization. A global average pooling operation aggregates the token representations, and a final *softmax* layer performs the binary classification.

**Mamba model:** We also evaluated state-space sequence models based on the recent Mamba architecture. Similar to the Transformer, the Mamba networks first apply a 1D convolutional patch embedding (*kernel size* = 32) to convert the raw PPG waveform into a sequence of tokens. The resulting feature map is transposed to a (*tokens*, *dmodel*) representation and passed through a single Mamba block (*depth* = 1), operating in the token space with model width (*dmodel*), state size (*dstate* = 64), local convolution width (*dconv* = 4), and *expansion factor* 2. These blocks implement a selective state-space model that jointly captures local and long-range temporal structure while maintaining linear complexity with respect to the sequence length. After the Mamba block, we perform global average pooling over the tokens, apply dropout, and use a final *softmax* layer for binary classification. We investigate two configurations. The first, "Mamba", uses *dmodel* = 64, non-overlapping patches (*stride* = 32), and 30% *dropout*, yielding a relatively higher-capacity model with fewer tokens. The second, "Mamba-v2", reduces the model width to *dmodel* = 48 but introduces 50% overlapping patches (*stride* = 16) and a slightly higher *dropout* rate (40%), approximately doubling the number of tokens and providing finer temporal resolution. These two variants enable us to examine the trade-off among model width, token density, and regularization in PPG-based affect recognition.

Due to the limited dataset size, increasing the model depth beyond a single block led to rapid overfitting for both the Transformer and Mamba architectures. We therefore adopt depth = 1 for all long-range models, yielding compact variants better suited to short PPG windows and small-sample training conditions.

*B. Data Preprocessing*

In this work, we follow the same preprocessing pipeline used in [22], applying identical filtering, segmentation, and normalization procedures to all four architectures to ensure consistency and reproducibility.

1. **Signal Filtering**: To enhance signal quality, we applied a third-order Butterworth bandpass filter with cutoff frequencies of 0.7–3.7 Hz, corresponding to heart rates of approximately 40–220 bpm. This range helps suppress motion artifacts and low/high-frequency noise. The filtering approach follows the methodology used in [35].

2. **Segmentation**: Following filtering, the PPG signal was divided into fixed-length windows to capture temporal variations relevant to emotional responses. We employed a sliding-window segmentation with 60-second windows and a 5-second overlap, consistent with previous studies [22][16]. Each segment was assigned an arousal, valence, and relaxation label based on the annotation protocol described in Section IV-A.

3. **Standardization**: Before model training, each segment was standardized using *Z-score* normalization.

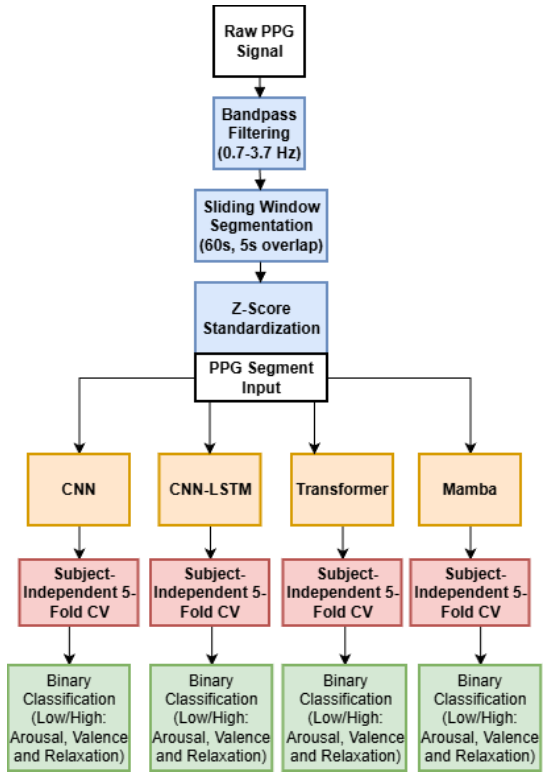

Figure 1 - Workflow of the Machine Learning Validation Pipeline for PPG-based Affect Recognition

IV. EXPERIMENTAL EVALUATION

*A. Dataset*

Affect representations capture more nuanced physiological variations and avoid limitations of categorical labels, which may not translate consistently across contexts or languages [36]. Accordingly, we chose the WARM-VR dataset [20], which provides valence, arousal, and relaxation annotations.

The WARM-VR dataset is a publicly available multimodal dataset for affect recognition in immersive, multisensory environments using wearable sensing instrumentation [37]. It was collected from 28 participants aged 19–37. The recording protocol involved inducing two affective states, stress, and relaxation, in addition to the baseline. Stress was induced via an arithmetic task, followed by relaxation through an immersive Virtual-Reality (VR) beach environment. The VR relaxation sessions incorporated synchronized multimedia stimuli: visual, auditory, and olfactory, with a crossover design for olfactory exposure (scented vs. scent-free). Participants wore a Polar H10 chest strap to record ECG signals, and an Empatica E4 wristband on their non-dominant hand, recording PPG at a sampling rate of 64 Hz. Each experimental phase (Baseline, Stress, Relaxation) lasted six minutes. Affective states were annotated subjectively using validated self-report questionnaires, specifically the Self-Assessment Manikin (SAM) for valence and arousal, and the Relaxation Rating Scale (RRS) for relaxation. According to the description of the WARM-VR dataset, data recorded from each participant included two VR relaxation sessions but only one stress-induction session; therefore, to avoid class imbalance, only the olfactory-enhanced VR relaxation session was used for the relaxation labels. This dataset is particularly relevant for wearable affective computing because it includes more participants than PPGE[13] or WESAD[14] datasets and uses wrist-worn PPG rather than fingertip sensors, as in DEAP[23].

*B. Evaluation Metrics*

We evaluate model performance using validation accuracy, the class-wise F1 scores (F1-0 and F1-1), and the average F1 score. Accuracy provides a general measure of correctness but may not adequately reflect performance differences between classes, particularly in affect-related tasks where class distributions may be uneven. This is because it cannot measure misclassification costs, which is a major drawback in real-world applications where these costs are often unequal [38]. To capture class-specific behavior, we report F1-0 and F1-1, which correspond to the F1 scores for the low and high affect states, respectively, (e.g., low vs. high arousal or low vs. high valence). The average F1 score is computed as the unweighted average of F1-0 and F1-1, giving equal weight to both classes and providing a more reliable measure than Accuracy under potential class imbalance.

*C. Model Training and Evaluation*

To prevent hyperparameter tuning from biasing the final evaluation on WARM-VR, we followed a two-stage training procedure. In the first stage, a trial-and-error hyperparameter tuning was performed on *learning rate*, *dropout*, and architectural width parameters, such as *dmodel*, using the PPGE dataset, which served exclusively as a development set. The second stage was evaluating the models on WARM-VR, where we adopted a subject-independent 5-fold cross-validation protocol to ensure that no subject's data appeared in both training and testing splits. This setup reflects real-world deployment scenarios in which models must generalize to unseen users. All reported results in this paper are based solely on the WARM-VR dataset, using the selected hyperparameters without further modification.

- **Training Configuration:** Following the training protocol described in [22], all models were trained with a batch size of 512, up to 350 epochs, and an early stopping patience of 80 epochs based on validation accuracy. This configuration was found to provide stable convergence across all architectures.

- **Class Weighting:** To address class imbalance across affective states, we used dynamic class weighting during training. The class weights were computed from the training fold at each split and applied to the loss function, ensuring that minority classes contributed proportionally to the gradient updates. This approach improved the model's sensitivity to underrepresented affect states.

- **Optimization and Loss Function:** For binary classification tasks (low vs. high arousal), we used categorical cross-entropy described above, with class weights, consistent across CNN, CNN–LSTM, Transformer, and Mamba models. The feed-forward networks within the Transformer blocks used GELU activation, whereas convolutional layers in CNN and CNN–LSTM models employed ReLU activation, matching the architectural specifications described in Section III. The Mamba models were implemented using the official PyTorch Mamba library, ensuring correct usage of selective state space modules and compatibility with the latest Mamba kernel implementations.

V. RESULTS AND DISCUSSION

Tables I summarize the classification performance of all evaluated architectures across the three affective states: Arousal, Valence, and Relaxation. Performance is reported using validation accuracy, F1-0, F1-1, and average F1, computed under a subject-independent 5-fold cross-validation protocol on the WARM-VR dataset.

*A. Results*

*1) Arousal Classification*

The CNN baseline achieved the highest validation accuracy (0.70), but the Transformer model ($d\_model$=64, $heads$=2) achieved the highest average-F1 (0.59), primarily by improving the low-arousal class (F1-0 = 0.54 compared to 0.47 for the CNN), at the cost of lower accuracy (0.63). Mamba models showed slightly lower average-F1 scores (0.57) and validation accuracy of 0.64 and 0.68 respectively. They did not surpass the CNN baseline, but performed better than the CNN-LSTM model (Val-accuracy 0.61 and average F1 0.52).

*2) Valence Classification*

The CNN baseline clearly dominates the other architectures. It achieved the strongest performance, with the highest accuracy (0.69) and the best average F1 score (0.63). Transformer variants performed moderately well, with close accuracy of 0.68 but a lower average-F1 value of 0.59. Mamba and Mamba_v2 had the lowest average F1 scores (0.52 and 0.53), indicating weaker discrimination capability of the valence dimension.

*3) Relaxation Classification*

For Relaxation, the CNN baseline achieved the highest accuracy (0.71) and an average F1 score 0.63 (F1-0 = 0.46, F1-1 = 0.79). The Transformer model obtained a slightly lower accuracy (0.68) but reached the highest average F1 (0.64), similar to the Arousal state, this was driven by an improved low-relaxation F1-0 of 0.52 while maintaining a high F1-1 of 0.76 . Transformer_v2 and both Mamba variants reached competitive performance (average F1 = 0.63). CNN-LSTM underperformed relative to all other models on this task.

*B. Discussion*

Across all three affective dimensions (arousal, valence, and relaxation), the results consistently show that the CNN baseline provides the highest validation accuracy for all tasks. Its average F1 scores were also either superior or within one percentage point of the best-performing model. In contrast, the CNN–LSTM architecture consistently underperformed, indicating that in our experiment, the temporal structure present in 60-second PPG segments does not provide a substantial advantage for recurrent models under the dataset's size and noise characteristics. The addition of recurrent layers may also introduce unnecessary complexity that is difficult to optimize effectively with limited data. One notable trend, however, is that

TABLE I. AVERAGE RESULTS OF 5-FOLD SUBJECT INDEPENDENT CROSS-VALIDATION ON WARM-VR

| Model | Val Accuracy | F1-0 | F1-1 | Average F1 |
|---|---|---|---|---|
| *Arousal only* | | | | |
| CNN | **0.70** | 0.47 | **0.68** | 0.58 |
| CNN-LSTM | 0.61 | 0.48 | 0.55 | 0.52 |
| Transformer | 0.62 | **0.54** | 0.63 | **0.59** |
| Transformer-v2 | 0.63 | 0.50 | 0.65 | 0.58 |
| Mamba | 0.64 | 0.50 | 0.63 | 0.57 |
| Mamba-v2 | 0.68 | 0.53 | 0.61 | 0.57 |
| *Valence only* | | | | |
| CNN | **0.69** | **0.50** | 0.76 | **0.63** |
| CNN-LSTM | 0.67 | 0.35 | **0.78** | 0.57 |
| Transformer | 0.68 | 0.40 | 0.77 | 0.59 |
| Transformer-v2 | 0.68 | 0.40 | **0.78** | 0.59 |
| Mamba | 0.63 | 0.30 | 0.74 | 0.52 |
| Mamba-v2 | 0.67 | 0.29 | 0.77 | 0.53 |
| *Relaxation only* | | | | |
| CNN | **0.71** | 0.46 | **0.79** | 0.63 |
| CNN-LSTM | 0.62 | 0.48 | 0.69 | 0.59 |
| Transformer | 0.68 | **0.52** | 0.76 | **0.64** |
| Transformer-v2 | 0.67 | 0.51 | 0.74 | 0.63 |
| Mamba | 0.67 | 0.49 | 0.76 | 0.63 |
| Mamba-v2 | 0.69 | 0.58 | 0.78 | 0.63 |

Transformer models produced more balanced F1-scores between the minority and majority classes. For both arousal and relaxation, the Transformer improved the F1-score of the low-affect class without substantially sacrificing the high-affect class. For example, in arousal classification, the Transformer increased F1-0 from 0.47 to 0.54 compared to CNN, while maintaining an F1-1 close to the CNN's value. A similar pattern appears in relaxation. The Transformer with *dmodel*=64 and *heads*=2 had the highest average F1 score for both arousal and relaxation states. This suggests that Transformer architectures may be less biased toward dominant classes and could be advantageous in scenarios where minority-class sensitivity is a primary objective. Although this did not translate into a higher accuracy, the improved balance in class-wise performance may be meaningful for certain application contexts. Mamba models demonstrated stable but not superior performance. While competitive in several cases, and generally stronger than the CNN-LSTM model, they did not outperform either the CNN or Transformer models in any affective dimension. This is likely because their long-range modeling capacity is underutilized in small, noisy, short-context PPG datasets.

Overall, the findings indicate that neither Transformer nor Mamba architectures currently provide clear advantages over simpler CNN models for contact PPG-based affect recognition, particularly when dataset size is limited. Combined with the complexity analysis in Table II, obtained by measuring inference time for a 60-s window on a Google Colab runtime with an NVIDIA L4 GPU (identical setting for all models), the results show that all architectures run in under 1.5 ms and that Mamba achieves slightly lower latency at the cost of substantially more parameters. Taken together, these findings suggest that CNNs remain an attractive choice due to their simplicity, compactness, and strong competitive performance; in the present setting, more complex sequence models do not offer sufficiently clear benefits to justify their added architectural complexity. At the same time, the evaluation is constrained by the small dataset size, short window lengths, and the absence of self-supervised pretraining, which may prevent long-range models from realizing their full potential. Until larger PPG datasets emerge and until pretrained models become more feasible, it will be important to revisit these comparisons to determine whether long-range sequence models can offer advantages under different data conditions.

TABLE 2. MODEL COMPLEXITY AND LATENCY FOR A 60-S PPG WINDOW

| Model | Nb of trainable parameter | Inference Time |
|---|---|---|
| CNN | 6,793 | 0.97 ms |
| CNN-LSTM | 6,098 | 1.50 ms |
| Transformer | 9,666 | 1.26 ms |
| Transformer-v2 | 35,649 | 1.24 ms |
| Mamba | 53,314 | 0.90 ms |
| Mamba-v2 | 35,186 | 0.87 ms |

VI. CONCLUSION

This study presented the first comparison of CNN, CNN-LSTM, Transformer, and Mamba architectures for contact PPG-based affect recognition using the WARM-VR dataset. Despite the widespread success of long-range sequence models in natural language and other time-series domains, our findings show that neither the Transformer nor the Mamba architecture offers a clear performance advantage over simpler CNN models for arousal, valence, or relaxation classification. CNNs consistently achieved the highest accuracy and competitive Average-F1 scores across all tasks, suggesting that the emotion-relevant information in wrist-based PPG is dominated by local features rather than long-range dependencies. Transformers provided more balanced class-wise performance, particularly for minority affective states, pointing to potential future value when class imbalance mitigation is a priority. Mamba models performed reliably but did not surpass Transformers or CNNs. Overall, these results highlight the practical reality that, given the limited size and noise characteristics of current wearable datasets, classical convolutional models remain the most effective choice for PPG-based emotion recognition. Future work should explore whether these findings hold across larger datasets or with pretraining approaches tailored to physiological signals.